**Lay Text Summarisation Using Natural Language Processing: A Narrative Literature Review.**

Oliver Vinzelberg[1], Mark David Jenkins[1], Gordon Morison[2], David McMinn[4] and Zoë Tieges[1,3,5*]

[1]School of Computing, Engineering and Built Environment, Glasgow Caledonian University, 70 Cowcaddens Road, Glasgow G4 0BA, Scotland, UK; oliver.vinzelberg@gmail.com; mark.jenkins@gcu.ac.uk; zoe.tieges@gcu.ac.uk.

[2]School of Computing, Engineering and Physical Sciences, University of the West of Scotland, High St, Paisley PA1 2BE, Scotland, UK; gordon.morison@uws.ac.uk.

[3]SMART Technology Centre, Glasgow Caledonian University, Glasgow, Scotland, UK.

[4]Lay Summaries Ltd, East Kilbride, Scotland, UK; davidmcminn@laysummaries.com

[5]Ageing and Health, Usher Institute, University of Edinburgh, Scotland, UK.

*Correspondence: zoe.tieges@gcu.ac.uk

**ABSTRACT**

Summarisation of research results in plain language is crucial for promoting public understanding of research findings. The use of Natural Language Processing to generate lay summaries has the potential to relieve researchers' workload and bridge the gap between science and society. The aim of this narrative literature review is to describe and compare the different text summarisation approaches used to generate lay summaries. We searched the databases Web of Science, Google Scholar, IEEE Xplore, Association for Computing Machinery Digital Library and arXiv for articles published until 6 May 2022. We included original studies on automatic text summarisation methods to generate lay summaries. We screened 82 articles and included eight relevant papers published between 2020 and 2021, all using the same dataset. The results show that transformer-based methods such as Bidirectional Encoder Representations from Transformers (BERT) and Pre-training with Extracted Gap-sentences for Abstractive Summarization (PEGASUS) dominate the landscape of lay text summarisation, with all but one study using these methods. A combination of extractive and abstractive summarisation methods in a hybrid approach was found to be most effective. Furthermore, pre-processing approaches to input text (e.g. applying extractive summarisation) or determining which sections of a text to include, appear critical. Evaluation metrics such as Recall-Oriented Understudy for Gisting Evaluation (ROUGE) were used, which do not consider readability. To conclude, automatic lay text summarisation is under-explored. Future research should consider long document lay text summarisation, including clinical trial reports, and the development of evaluation metrics that consider readability of the lay summary.

# INTRODUCTION

Natural Language Processing (NLP) is the discipline concerned with how computers can understand, manipulate, or summarise human language in forms of text or speech (see Appendix A: Table A1 for a list of abbreviations and acronyms). A key focus lies in a model's ability to capture important information in language using machine learning and rule-based language modelling [1]. Selecting and retaining important parts of an input sequence and understanding context remains a challenge for machines [2].

Automatic text summarisation, a subfield of NLP, involves creating a concise and accurate overview of a text document whilst preserving the critical content and the overall context. Automatic text summarisation has been mostly oriented towards news or science articles, due to the high availability of human annotated data sets [3]. Given the exponential increase in the amount of research published every year, there is an overwhelming amount of scientific information available. Automatic text summarisation has the potential to help reduce this information overload, potentially leading to time and cost savings [4]. Concise summaries make the exponential growth of scientific publications each year more manageable to read [5]. Indeed, there has been growing interest in automatic text summarisation as a method for helping researchers, clinicians and other stakeholders seeking information to efficiently obtain the "gist" in a given topic by producing a textual summary from one or multiple document [6].

A common distinction in text summarisation methods is between extractive and abstractive approaches [7, 8]. In extractive text summarisation, phrases of the original text are ranked according to their importance, reordered, and used to create the summary. The ranking is done by assigning relative weights based on the frequency of a word or phrase, or by comparing it to a corpus and determining whether a phrase belongs to a particular domain.

Due to the rise in deep learning methods in recent years, the focus has shifted toward abstractive summarisation [9]. Abstractive text summarisation is more closely aligned with human-created summaries. In contrast to extractive summarisation, abstractive summarisation generates words and phrases that are not present in the original text which are then used for the summary. This method is more challenging than extractive summarisation methods because it does not guarantee a baseline in terms of grammar quality or accuracy of information. Recent advances have made it possible to generate longer abstract summaries using pre-trained transformer models [10]. A comprehensive review of text summarisation methods has been published previously, including detailed discussion on extractive and abstractive approaches [11], as well as their application to the summarisation of medical content [12].

Lay text summarisation is a subfield of automatic text summarisation, concerned with generating summaries of text in lay language (i.e. summaries that are understandable to a non-expert audience). This supports the greater goal of making research findings more accessible to the wider public, particularly to people unfamiliar with academic writing, and people of all age groups. Plain language summarisation presents specific challenges, including the need to provide relevant background information for certain concepts, clarification of terminology, and the use of simple sentence structures; aspects that are not necessary for general text summarisation [13]. Extractive summarisation approaches may not be as suitable for lay text summarisation because the source text should not only be summarised, but also simplified.

While there are review articles on research and news article summaries (e.g. [14]), none focus on lay text summarisation. Our review therefore seeks to give an overview and narrative synthesis of the literature on state-of-the-art automatic text summarisation approaches for the purpose of generating lay summaries.

While this review focuses on NLP methods for text summarisation, especially transformer models, we do not thoroughly discuss the underlying models. Below, we provide a brief introduction to transformer models and evaluation metrics for NLP. We refer the interested reader to Wolf et al. [15] and Appendix B for further theoretical background on NLP and transformer models.

**Transformer models for NLP**

Transformers have rapidly become the standard architecture for NLP, surpassing convolutional and recurrent neural networks in performance for both natural language understanding and generation tasks. Transformers allow parallel training with high efficiency, capturing long-range sequence features, and scaling according to training data and model size [15]. Transformers are non-recurrent sequence-to-sequence tools that combine an encoder with a decoder and usually consist of attention mechanisms to improve context capturing [16]. In principle, self-attention maps the relationships between words at different positions in an input sequence to create a representation and capture the relative importance of tokens compared to others [17]. In the encoder, self-attention calculates an attention score based on the output of the previous layer and compares the score with all other scores of words of the input sequence. The advantages of transformers include easier parallelisability and shorter training time compared with recurrent or convolutional approaches. Self-attention layers are usually stacked multiple times in the encoder and decoder to improve the computation of dependency relationships between words of the same sentence.

Transformers have opened a new chapter in NLP. Some authors go as far as to call them a revolution [18, 19]. To illustrate, pre-trained transformer models (e.g. Open AI Transformer) were able to achieve 10-20% better results than the previous state-of-the-art models in various NLP challenges, such as SQuAD (Stanford Question Answering Dataset) [20, 21], SNLI (Stanford Natural

Language Inference Corpus) [22], SRL (Semantic Role Labelling) and SST-5 (Stanford Sentiment Treebank [23]. Transformers are pre-trained on large data sets and can be fine-tuned to solve a variety of NLP tasks, similar to computer vision models which are pre-trained on ImageNet and then fine-tuned to work on different computer vision tasks [24] [25].

**Evaluation metrics**

Metrics are important for model evaluations and comparisons. Popular metrics in machine learning include accuracy, recall, precision, and F1-Score (the harmonic mean of precision and recall). Yet, in text summarisation, the choice of metrics is not straightforward because generated words, phrases, and whole sentences need to be evaluated instead of labels. One approach is to have humans, ideally subject matter experts, evaluate the generated summaries based on readability, coherence, grammar, and information value. However, this is time-consuming, which renders this approach unfeasible for most cases since developers rely on quick performance evaluation of their models [13].

Commonly used evaluation metrics for text summarisation models are Recall-Oriented Understudy for Gisting Evaluation (ROUGE [26, 27]), Bi-Lingual Evaluation Understudy (BLEU [28]), and Metric for Evaluation of Translation with Explicit ORdering (METEOR [29]). ROUGE measures the lexical overlaps between generated summaries and the corresponding full texts. Bi-Lingual Evaluation Understudy (BLEU) is a precision focused metric for evaluating generated texts, which is mainly used for evaluating machine translation tasks [28]. This can also be applied to text summarisation as it compares the closeness of a generated text to a reference text by matching n-grams (i.e., a sequence of N words). METEOR calculates the harmonic mean (F1) of unigram precision and recall, while weighting recall higher than precision. Developing metrics for NLP tasks is an ongoing research area, and suitability varies greatly depending on the task and dataset.

**Aims of the study**

The aim of this narrative literature review is to describe and compare the different text summarisation approaches used to generate lay summaries. As a background to this literature review stands the recently proposed EU Regulation No. 536/2014 [30], which requires pharmaceutical companies to publish a summary in lay terms alongside the full clinical trial report. Its goal is to make clinical trial reports more accessible to the general public with a view to empowering them to make more informed health-related decisions. The Clinical Trials Expert Group (CTEG) [31] argue that lay summaries should be written in plain and clear language, accessible for people from the age of 12 upwards, and be as short as possible. Moreover, it must cover the main objectives and findings of a study while using everyday conversational language and avoid statistical terms or scientific language.

The following research questions are addressed:

- Which text summarisation methods using NLP have been applied to the domain of lay text summarisation?
- How is the performance of models for lay text summarisation assessed?
- Which lay text summarisation approaches are the most effective?
- What are the current challenges and future outlooks of research in this area?

**MATERIALS AND METHODS**

This narrative literature review provides an overview and narrative synthesis of the existing knowledge on automated lay text summarisation using NLP, based on available published research. While there are no acknowledged reporting guidelines for narrative reviews, we followed best practice recommendations, including clear inclusion and exclusion criteria for literature search and a critical evaluation of selected articles (e.g., regarding key results, limitations, interpretation of results and the impact of conclusions on the field) [32]. Relevant papers were identified using comprehensive electronic searches of Google Scholar, Web of Science, IEEE Xplore, Association for Computing Machinery (ACM) Digital Library and arXiv for articles published from inception until 6 May 2022, using a combination of search terms for (lay) text summarisation of clinical trial reports or research articles with automatic text summarisation methods as follows: (lay summarisation OR lay sum OR lay term summarisation OR automatic text summarisation OR clinical trial reports OR research articles) AND (nlp OR machine learning OR deep learning OR transformers OR nlp metrics). Bibliographies of relevant papers and books were scanned for further articles of potential relevance.

Studies were included if they presented original work concerned with automatic text summarisation methods addressing the task of lay summarisation. Studies were excluded if they were not relevant to the topic of automatic generation of lay text summaries, or if they did not cover NLP techniques relevant to lay text summarisation.

One study author (O.V.) collected the relevant literature and assessed each paper against the inclusion and exclusion criteria to determine whether it should be included in the review. Papers for which a clear decision could not be made were discussed with another study author (Z.T.) and a consensus reached. Information was extracted from each paper on pre-processing, methodology including data sources and models, training and hyperparameters, main findings and conclusions, choice of evaluation metrics and model performance.

## RESULTS

**Study identification**

The search yielded 82 articles which subsequently underwent full-text review to assess eligibility. This resulted in a final selection of eight papers that were included in the narrative synthesis [3, 7, 10, 33-37]. The main reasons for exclusion were a lack of addressing lay text summarisation or focusing on other NLP areas such as text translation rather than summarisation. The included articles were published in 2020 and 2021 and were all linked to the Scholarly Document Processing workshop 2020 [7] at the 2020 Conference on Empirical Methods in Natural Language Processing (EMNLP 2020), which proposed three tasks: CL-SciSumm (Scientific Research Summarisation), LaySumm (Lay Summarisation) and LongSumm (Long Scientific Document Summarisation) to improve the state-of-the-art for different aspects of scientific document summarisation. For the LaySumm task, a corpus comprising 572 author-generated lay summaries from a multidisciplinary series of journals in Materials Science, Archaeology, Hepatology and Artificial intelligence, along with their corresponding abstracts and full text articles, was provided by Elsevier. Yu et al. [34] note that there were a few outliers not containing an abstract or introduction. An example of the data set is available on GitHub [38]. According to the organisers of this challenge, LaySumm should aim to give a succinct summary spanning around 100 words stating the aim and results of the research and its potential impact whilst omitting technical jargon [7]. In addition, it should be coherent in its structure and semantics, and interesting to read. The lay summary should cover all the main points of an article whilst not being redundant.

Of the eight included papers, one paper was published by the organisers of the challenge [7], with the others being published by participants proposing their methods [3, 10, 33-37]. Two articles were published on arXiv [10, 34], and the other six articles were published in Proceedings of the First Workshop on Scholarly Document Processing [3, 7, 33, 35-37].

In the following sections, we describe and compare the selected studies, focusing on the study methodologies, including use of extractive/abstractive approaches, use of transformer models, input data format/structure, model training strategies, and data augmentation techniques. To conclude, a number of recommendations are proposed based on the findings from the review.

**Automatic text summarisation methods for lay summaries**

Table 1 provides an overview of study characteristics: models, methodologies, input data, training and results (ROUGE scores) of the CL-LaySumm 2020 challenge. The table is ranked from best to worst ROUGE scores. Table 2 provides an overview of the main study findings and conclusions.

A combination of extractive and abstractive summarisation methods in a hybrid approach was found to be the most successful strategy to generate lay summaries. DimSum [34], the winners of the competition according to ROUGE scores, combined a Pre-training with Extracted Gap-sentences for Abstractive Summarization (PEGASUS) model [39] for abstractive text summarisation in addition to a Bidirectional Encoder Representations from Transformers (BERT) model for extractive text summarisation [3]. BERT is a language representation model that is pre-trained on large unlabelled text data which allows fine tuning on smaller data sets for a wide range of downstream tasks. Context is encoded bidirectionally by masking words of the input data, thereby resolving the issue of earlier transformers that could only encode unidirectionally [40]. On the other hand, PEGASUS is tailored for the task of abstractive text summarisation. It is pre-trained by using Gap-Sentence Generation, and similar to BERT, works well on small data sets [39].

The PEGASUS transformer was used to generate most of the lay summaries. However, when the lengths of the produced summaries were too short, a BERT transformer model was used in an extractive manner to identify important sentences to add if they met an author-defined readability metric [3].

Four of the seven participating groups addressed this challenge by combining elements of extractive and abstractive summarisation methods in a hybrid approach [3, 10, 34, 36] (Table 1). Roy et al. [10] experimented with a hybrid approach, using SummaRuNNer [41] pre-trained on the PubMed data set for extractive summarisation as well as Bidirectional Auto-Regressive Transformers BART and T5 to generate abstractive summaries. However, compared with the hybrid approach, they were able to achieve better results using a purely abstractive approach based on BART. Similarly, Seungwon [3] generated summaries using PreSumm (which uses BERT) for extractive summarisation in combination with PEGASUS for abstractive summarisation. Chaturvedi et al. [36] extracted sentences from the introduction, discussion and conclusion sections using BioBERT [42]. The abstract and extracted sentences were summarised separately in an abstractive manner, using BART, and then concatenated. These groups all achieved higher ROUGE extracted sentences were summarised separately in an abstractive manner, using BART, and then concatenated. These groups all achieved higher ROUGE scores compared with the groups that focused on a single summarisation method (either extractive or abstractive). This suggests that a hybrid approach may be key to a successful lay summarisation process (Table 2).

Transformer-based methods such as BERT or PEGASUS were the most commonly used models, with six of the seven groups utilising these methods [3, 10, 34-37]. Four groups used Bidirectional Auto-Regressive Transformers (BART) [10, 34, 36, 37], two groups used PEGASUS [3, 35], whereas SummaRuNNer [34], Presumm (which requires a pre-trained BERT model [3]) and BioBERT [36] were only used once each (Table 1).

Yu [34], Roy [10], Chaturvedi et al. [36] and Mishra [37] followed an abstractive summarisation approach using BART. BART is a similar large pre-trained transformer to BERT, however, BART follows a different pre-training method by reconstructing a text that was corrupted using a noise function [47], whereas BERT is pre-trained using Masked Language Modelling (MLM) and Next Sentence Prediction (NSP) [40]. Mishra et al. [37] achieved mixed results with their abstractive approach, whereas the other groups achieved better performance scores because they used a hybrid approach combining their abstractive approach with an extractive step.

Participants used transformers pre-trained on different data sets, such as CNN/DM, PubMed, PMC and arXiv articles. Transformer models pre-trained on the CNN/DM news corpus were among the top performers. Two groups [34] [3] used BART models pre-trained on the CNN/ DM data set which includes over one million news articles and their summaries. This seems to indicate that pre-training on diverse general domain data is effective, even in the context of scientific lay text summarisation. Yet, Gu et al. [44] showed that pre-training BERT on domain-specific data may be advantageous since the model can operate on in-domain vocabulary. Indeed, several domain-specific pre-trained BERT models have been proposed in recent years, for example BioBERT, pre-trained on the PubMed data set [42], SciBERT, pre-trained on research areas of computer science and biomedicine [45], BERTweet, pe-trained on tweets [52], Med-BERT, pre-trained on electronic health records [46], and FinBERT, pre-trained on financial data [47]. However, in this challenge only one group [36] made use of a domain-specific transformer (BioBERT) though they were not among the top performers according to ROUGE scores (Table 1).

**Input data**

With regards to input data, Cohan et al. [48] show that section-level processing of scientific documents is an effective approach to improve summarisation results. Further, Collins [49] conclude that not all sections are equally useful. Recent research has shown that a hierarchical or hybrid approach to summarising scientific documents, whereby an extractive summary of each section is produced independently at the first level and the sectional output is then abstracted into a brief summary at the second level, is highly effective [50]. This concept could also be applied to lay summarisation. In fact, two of the groups that participated in the CL-LaySumm 2020 challenge reported that not all sections of a paper were equally important for creating lay summaries [10, 34] (Table 2). Most groups only used particular sections of a text (Abstract, Introduction, Conclusion, and/or Discussion) as input data, which improved the model performance especially when the Abstract was included. All of the groups, apart from Mishra et al. [37], reported using the Abstract (or a combination of the Abstract with extracted sentences of the full text) as the sole input to generate their lay summaries. This decision was informed by experimenting with using different sections of the input data, as well as the limited token size of the available transformers (Table 1).

**Model training**

With regards to training the model, a training data set split of around 80-90% was common among the top performers. The three poorest performing groups did not provide details of their data split ratio. Of note, the two top performing models reported by Yu et al. [34] and Seungwon [3] differ in the number of training steps used, namely 6,000 versus 20,000 to 50,000, respectively. Both used a batch size (i.e., the number of training examples used by one GPU in one training step) of 1 due to the limited available computing power.

On the other hand, no consistent pattern of data pre-processing was evident. In general, the top performing groups provide a more detailed overview of hyperparameters, methodology, and pre-processing compared with the other groups.

Although the lowest-performing group [37] did not elaborate on their methodologies, based on the information provided it can be assumed that they were the only group not utilising a transformer-based model. Instead, they used the Maximum Marginal Relevance (MMR) metric to classify sentences according to their importance.

A common criticism reported by authors was the small data set size of only 572 sample articles [3, 34]. It is likely that a larger data set would have resulted in increased ROUGE scores and could have prevented overfitting.

**Data augmentation**

Finally, data augmentation involves artificially increasing the amount of data by manipulating or modifying existing data. Small data sets, such as the one used in the CL-LaySumm 2020 challenge, tend to cause overfitting, and data augmentation can provide a regularisation effect. Li [51] categorised data augmentation methods for NLP into three categories, namely methods based on paraphrase, noising, and sampling. Paraphrase-based methods are concerned with replacing words with their synonyms, or round-trip translating data, whereas noising-based methods are concerned with swapping, deleting, and substituting words. Sampling-based methods are based on language models and self-training to change sentence structure while maintaining grammatical correctness.

Only one group in the CL-LaySumm 2020 challenge tried to enhance the data set by a data augmentation method where words from the original corpus were replaced by synonyms [34]. However, this resulted in an increased level of noise rather than improved model performance.

Other studies have reported data augmentation methods for fine tuning and improving pretrained models for summarisation. For example, Fabbri et al. [52] used data augmentation to improve abstractive text summarisation models by performing round-trip translations on their data, which has been shown to be an effective way to paraphrase text while preserving semantic meaning. The authors found that this approach increased their models' performance. Thus, data augmentation can be an effective strategy to enrich a data set. More investigation in the area of data augmentation for lay summarisation is needed.

**Recommendations**

To conclude, we propose a number of recommendations for automatic lay text summarisation based on the reviewed literature:

- Hybrid approaches combining extractive with abstractive text summarisation methods may be important for successful lay text summarisation.
- Extractive summarisation applied to each section of an input text document can be useful to identify the most important sections for a summary (e.g. the Abstract, Introduction or Conclusion).
- Transformers are the state-of-the-art for lay text summarisation. The most appropriate transformer depends on the specific use case.

**DISCUSSION**

This literature review outlines the NLP methodologies that have been applied to the area of lay text summarisation. It is demonstrated that, while text summarisation is a widely researched topic in NLP, there is a surprising lack of research focusing on lay text summarisation. We were able to identify only 8 relevant papers, published between 2020 and 2021, highlighting the novelty of this research area. All papers were linked to the CL-LaySumm 2020 challenge from the Scholarly Document Processing workshop 2020.

**Successful approaches for lay text summarisation**

The evidence base, though small, suggests that lay text summarisation is best approached using a hybrid method that combines the advantages of both extractive and abstractive summarisation by first ranking sentences or paragraphs of a text in order of importance and then creating an abstractive summary of the extracted sentences. Moreover, the success of transformer models for text summarisation is well documented, and they are also the preferred models for lay text summarisation. However, it is not possible to conclude from the studies reviewed here which combination of transformers, hyperparameters, pre-training and fine tuning achieves the best results. Comparing the participants' approaches to the CL-LaySumm 2020 challenge showed that hyperparameters and pre-processing applications varied greatly and were not consistently reported. We tentatively conclude that the following steps may contribute to high model performance: pre-processing the data by applying sentence ranking, cleaning, dividing the text into different parts by prioritising local over global context or truncating the text according to the token size of the transformer; hyperparameter optimisation; and experimenting which sections to use as input. Lay text summarisation can further benefit from first applying extractive approaches to each section individually and then feeding the results into a transformer for abstractive summarisation. This approach is supported by Xiao and Carenini [50], who showed that local context is crucial to improve a summarisation method, whereas global context may be neglected.

     CL-LaySumm 2020 was the only data set for lay summaries found in the literature, however, it is not publicly available. Most research in text summarisation has relied on public data sets (such as WikiDes [53]) while only a small number of studies have used private data sets. This is supported by Widyassari et al. [54] who looked at 85 text summarisation studies, of which 55 operated on public data sets, most commonly Document Understanding Conferences (DUC) news data sets. Interestingly, as the field of text summarisation grows, so too do the amount of data sets, including benchmarks, which combine datasets with evaluation metrics and baselines. These benchmarks provide an integrated evaluation setup for comparing NLP systems systematically (e.g. CiteSum [5] and CiteBench [55]).

**Considerations for future research into lay text summarisation**

There is a clear need for further research into lay text summarisation to provide a wider audience with access to scientific results. The available studies on lay text summarisation only focused on short research articles. In future work, lay summarisation of longer documents should be explored, for example clinical trial reports. Similar to the EU Regulation No. 536/2014, which requires pharmaceutical companies to publish a lay summary of clinical trials, this type of 'mandated' provision of lay summaries could also be implemented across other fields. Such mandated lay summarisation would go a long way to help promote the dissemination, and potential uptake, of important scientific findings. Lay text summarisation based on NLP could help automate these processes, saving time and money for those responsible for their development [56]. Yet, automating this task remains a challenge, given that the source documents for these summaries often vary in structure and format, are typically lengthy, and the nature of the content can be complex and esoteric. As such, this area warrants further research.

One area requiring specific attention is the automatic text summarisation of long documents. This task is challenging because transformers cannot encode long sequences due to their self-attention mechanisms, which enable transformers to retain contextual information contained in a sequence. These self-attention steps scale quadratically with sequence length. In this respect, the effectiveness of the Longformer transformer, a linearly scaling self-attention mechanism-based transformer model ([57]), in generating lay summaries of long text documents should be explored. Similarly, Zaheer [58] proposed a linear self-attention mechanism that retains all functions of full self-attention mechanisms. Longformer was not utilised by the participants of the CL-LaySumm 2020 challenge, even though some groups reported struggling with the token size [34, 36].

Previous workarounds split documents into different sections, such as the divide and conquer approach by Chaturvedi et al. [36], or simply cut off sections of text, such as Roy et al. [10]. Notably, this approach may cause loss of important information. Similarly, Xiao and Carenini [50] had success by taking local and global contexts into account. However, they observed that large transformer models such as BERT performed poorly in such cases. Consequently, they opted for an alternative LSTM-based approach.

With regards to evaluation metrics, ROUGE was used in the reviewed studies to evaluate model performance. However this metric is limited in that it does not consider readability (i.e. how easily a passage can be understood by a reader) or semantic and factual accuracy [26, 27]. Indeed, there have been calls for new evaluation metrics needed in the area of NLP [59]. Ng and Abrecht [60] introduced ROUGE-WE, an extension to ROUGE, which considers the cosine similarity during matching. Similarly, BertScore compares generated and original text at the token level while also considering cosine similarity [61]. In addition, readability indices and human evaluation are also important to assess model performance. Readability evaluation metrics such as Flesch-Kincaid grade level [62] estimate the years of education generally required to understand the text. Yet, while these metrics are useful, they do not capture key aspects of the output text, such as fluency, grammaticality, style, and factual correctness. To consider these properties, methods for assessing the summary quality by human evaluators are needed [13, 63].

The lack of comprehensive up-to-date studies on evaluation metrics for text summarisation and a lack of consensus on evaluation protocols continue to hinder progress [64]. As such, there is a need for the development of metrics that more effectively evaluate the performance of lay summary models. Notably, other metrics such as COMET have shown promise in a summarisation context [65].

**Conclusions**

We reviewed the literature on state-of-the-art text summarisation approaches to generate lay text summaries. Our results show that (i) research on automatic lay text summarisation has received little attention to date; (ii) the most effective models are based on transformers; (iii) a hybrid approach combining abstractive and extractive text summarisation is recommended; and (iv) pre-processing the input texts, such as applying extractive summarisation or determining which sections of a text to include appear to be a major factor in the effectiveness of models.

To our knowledge, this is the first literature review on automatic lay text summarisation. Some limitations of this review should be acknowledged. We identified a relatively small number of studies, which do not allow definitive conclusions regarding the most effective approaches for lay text summarisation. Only one author (O.V.) performed the literature search. Further, all studies used the same dataset (not publicly available) and the same group of evaluation metrics (ROUGE), and all papers were linked to the CL-LaySumm 2020 challenge, which clearly limits generalisability of findings. Chandrasekaran et al. [7] summarise the methodologies of the studies from the CL-LaySumm 2020 challenge and provide a comparison between lay summaries and typical paper abstracts (Technical Summaries). Our literature review adds to the Chandrasekaran et al. paper by providing additional detail and critical discussion on the methods of the selected studies (including pre-processing, model training and data augmentation). Further, our review highlights the methodological aspects that may contribute to successful lay summarisation, thereby providing a starting point for further development and fine tuning.

The development of lay summaries of scientific findings across different fields would help to facilitate the dissemination and uptake of important discoveries, potentially having a positive impact on the broader society. Automatic lay text summarisation based on NLP techniques is a promising method to help automate these processes.

*Supplementary Materials*

Appendix A: Table A1, List of abbreviations and acronyms used in the literature review.

Appendix B: Theoretical background on Natural Language Processing and Transformer models.

*Author Contributions*

Conceptualisation, all authors; methodology, O.V., M.D.J., Z.T., and G.M.; investigation, O.V.; resources, M.D.J.; writing—original draft preparation, O.V. and Z.T.; writing—review and editing, O.V., Z.T., and M.D.J..; supervision, Z.T., M.D.J. and G.M.; project administration, O.V., Z.T., D.M., and G.M.; funding acquisition, D.M., Z.T. and G.M. All authors have read and agreed to the published version of the manuscript.

*Funding*

This research was funded by The Data Lab (Principal Investigator: Z.T.) and Lay Summaries Ltd.

*Informed Consent Statement*

Not applicable.

*Data Availability Statement*

Not applicable

*Conflicts of Interest*

O.V., M.D.J., Z.T., and G.M. declare no conflicts of interest. D.M. is a shareholder of Lay Summaries Ltd.

Table 1. Overview of the methodologies and results of the CL-LaySumm 2020 challenge. ROUGE (Recall-Oriented Understudy for Gisting Evaluation) was used as the main evaluation metric. ROUGE-1 compares the overlap of unigrams, ROUGE-2 calculates the overlap of bigrams, and ROUGE-L-F1 calculates the F1 score of the longest overlapping phrases.

| Model name, publication | Pre-processing | Model/ Algorithm | Methodology | Input text sections used. | Training | ROUGE |
|---|---|---|---|---|---|---|
| DimSum, Yu et al. 2020 [34] | Removed tags and outliers and removed input data samples that had no abstract or introduction. Truncated input texts to a maximum length of 1024 tokens. | BART and SummaRuNNer. | Hybrid approach. Abstractive summarisation (BART) combined with extractive model (SummaRuNNer) to maximise ROUGE. | Experimented with different combinations of the Abstract, Introduction and Conclusion. Best results achieved by only using the abstracts. | 90/10 split of data set. Hyperparameters: dynamic learning rate, warm up 1000 iterations. Batch size (i.e., the number of training examples used by one GPU in one training step) of 1 (due to GCU memory limitations). Training for 6000 iterations. | Rouge1-F1: 0.4600 Rouge2-F1: 0.2070 RougeL-F1: 0.2876 |
| Seungwon (2020) [3] | Not provided | PEGASUS and Presumm.(which uses BERT) | Hybrid approach. Abstractive summarisation (PEGASUS) combined with extractive summarisation (Presumm) to improve quality of produced summary. | Only the Abstracts used for abstractive summarisation. The entire text used for extractive summarisation but prioritising the Abstract (to generate additional sentences for the produced summary). | 80/10/10 split of data set. Hyperparameters: Approach 1: 20,000 steps training, batch size of 1 (due to GCU memory limitations), learning rate 0.0001. Approach 2: Pre-training for | Rouge1-F1: 0.4596 Rouge2-F1: 0.2146 RougeL-F1: 0.2977 |
| | | | | | 50,000, fine-tuning for 10,000 steps. | |
| Summaformers, Roy et al. 2021 [10] | Not provided | BART | Abstractive summarisation approach (BART). | Experimented with using different sections of the texts. Concluded that using Abstracts as the only section is most effective. | 80/20 split of data set. Hyperparameters: Adam optimiser, learning rate 0.00005, learn rate schedule based on ROUGE-1, repetition penalty of 1.8 in the hyperparameter tuning phase. | Rouge1-F1: 0.4594 Rouge2-F1: 0.1902 RougeL-F1: 0.2744 |
| AUTH, Gidiotis et al. [35] | Lowercased the text, replaced Greek symbols with distinct words, re-moved unwanted tokens, equations and references. | PEGASUS. | Abstractive summarisation (PEGASUS). | Solely used Abstracts to generate abstractive summaries. | 6/2/2 split of data set. Model fine-tuning. Hyperparameter details not provided. | Rouge1-F1: 0.4456 Rouge2-F1: 0.1936 RougeL-F1: 0.2772 |
| DUCS, Chaturvedi et al. 2020 [36] | Removed redundant white spaces, hyper-links, references and sentences containing more than 1/5 of special characters. | BART and BioBERT. | Hybrid approach. Abstractive summarisation (BART) combined with extractive summarisation (BioBERT). | Extractive step performed on the Introduction, Discussion and Conclusion sections. Abstractive summarisation performed on the extracted sentences combined with | Not provided. | Rouge1-F1: 0.4253 Rouge2-F1: 0.1748 RougeL-F1: 0.2526 |

| | Replaced common acronyms by their full meaning. Removed all punctuation except full stops, exclamation and question marks. | | | the Abstract, and abstractions systematically merged. | | |
|---|---|---|---|---|---|---|
| IIITBH-IITP, Reddy et al. 2020 [33] | Removed complex words. Lemmatised words and removed advanced symbols from texts. | Maximum Marginal Relevance (MMR). | Extractive summarisation approach using the MMR ranking algorithm. | Experimented with using the full text, Abstract and Conclusion. Best results achieved by only using the Abstract. | Not provided. | Rouge1-F1:0.4048 Rouge2-F1: 0.1690 RougeL-F1:0.2244 |
| IITP-AI-NLP-ML, Mishra et al. 2020 [37] | *Not provided.* | BART. | Abstractive summarisation approach (BART). | *Not provided.* | Two Adam optimisers with $\beta1 = 0.9, \beta2 = 0.99$. For the encoder: learning rate of 0.002 and 20,000 warm-up steps. For the decoder: learning rate of 0.1 and 10,000 warm-up steps. *Data set split not provided.* | Rouge1-F1: 0.3132 Rouge2-F1: 0.0631 RougeL-F1:0.1662 |

Note: Recall-Oriented Understudy for Gisting Evaluation (ROUGE; score range 0-1, higher scores indicate better model performance), Bidirectional Encoder Representations from Transformers (BERT), Pre-training with Extracted Gap-sentences for Abstractive Summarization (PEGASUS), Term Frequency-Inverse Document Frequency (TF-IDF), Text Summarisation with Pretrained Encoders (PreSumm) which is a framework based on BERT that also allows for abstractive summarisation [43].

Table 2 Overview of the main study findings and conclusions of the included papers.

| Model name, publication | Main study findings and conclusions |
|---|---|
| DimSum, Yu et al. 2020 [34] | Experimented with using different section combinations of texts as input data.<br>Explored data augmentation by replacing words with their synonyms in texts.<br>Data set too small, data augmentation not helpful, adding CLS (classification) tokens to the beginning of sentences increased precision. |
| Seungwon (2020) [3] | Calculated ROUGE scores of lay summaries and Abstracts/ full texts of the data set.<br>Abstracts were selected as input data for abstractive summarisation. The entire text was used for extractive summarisation but prioritising the Abstract (to generate additional sentences for the produced summary).<br>A hybrid approach, combining abstractive and extractive text summarisation, was found to be most effective. compared to either extractive or abstractive summarisation methods alone.<br>Without the concatenation of extracted sentences, summaries would be too short in length.<br>Readability metric was included to evaluate quality of the output which improved the performance slightly. |
| Summaformers, Roy et al. 2021 [10] | Using the Abstract alone as input source to generate lay summaries, compared to using Abstracts combined with other sections, achieved the best results.<br>Sequence lengths of 140 for BART resulted in the best performance. |
| AUTH, Gidiotis et al. [35] | The PEGASUS model pretrained on the PubMed data set (compared to the pre-trained PEGASUS model and a model fine-tuned on the arXiv dataset) performed the best, especially regarding ROUGE-2 and ROUGE-L scores.<br>The model pre-trained on the arXiv data set also performed well. |
| DUCS, Chaturvedi et al. 2020 [36] | Following extractive text summarisation, the abstractive summarisation step improved the performance significantly.<br>Dividing the text into segments (Abstract, Conclusion, Introduction, and Discussion sections) as input source for extractive summarisation, is most useful for generating lay summaries. Model performance improved when creating extractive summaries for individual sections compared to the entire text. |

| | Using Weighted Minimum Vertex Cover (wMVC) further improved ROUGE scores. |
|---|---|
| IIITBH-IITP, Reddy et al. 2020 [33] | Performance increased through consideration of the Maximum Marginal Relevance (MMR) metric and using Abstracts to generate summaries. A word removal approach using a lexical database like WordNet failed due to non-presence of scientific terms. The need for a more sophisticated approach is acknowledged. |
| IITP-AI-NLP-ML, Mishra et al. 2020 [37] | Not provided |

Note: Recall-Oriented Understudy for Gisting Evaluation (ROUGE; score range 0-1, higher scores indicate better model performance), Pre-training with Extracted Gap-sentences for Abstractive Summarization (PEGASUS), Term Frequency-Inverse Document Frequency (TF-IDF).

**Appendix A** List of abbreviations and acronyms used in the literature review.

**Table A1.** Abbreviations and acronyms used in the literature review.

| Abbreviation/ Acronym | Meaning |
| --- | --- |
| LaySumm | Lay summarisation |
| BART | Bidirectional Auto-Regressive Transformers |
| BERT | Bidirectional Encoder Representations from Transformers |
| BLEU | Bilingual Evaluation Understudy |
| CLS | Special Classification Token |
| CNN | Convolutional Neural Network |
| CNN - DM | Cable News Network - Daily Mail |
| CTEG | Expert Group on Clinical Trials |
| DUC | Document Understanding Conference(s) |
| ELMo | Embeddings from Language Model |
| GPU | Graphics Processing Unit |
| GRU | Gated Recurrent Units |
| LSTM | Long Short-Term Memory |
| METEOR | Metric for Evaluation of Translation with Explicit ORdering |
| MLM | Masked Language Modelling |
| MMR | Maximal Marginal Relevance |
| NLP | Natural Language Processing |
| NSP | Next Sentence Prediction |
| PEGASUS | Pre-training with Extracted Gap-sentences for Abstractive Summarisation |
| RNN | Recurrent Neural Networks |
| ROUGE | Recall-Oriented Understudy for Gisting Evaluation |
| SVM | Support Vector Machines |
| TF-IDF | Term Frequency-Inverse Document Frequency |
| TPU | Tensor Processing Unit |
| ULMFiT | Universal Language Model Fine-Tuning |
| wMVC | Weighted Minimum Vertex Cover |

**Appendix B** Theoretical background on Natural Language Processing and Transformer models.

**Natural language processing**

Natural Language Processing (NLP; see Appendix A: Table A1 for a list of abbreviations and acronyms) is the discipline concerned with how computers can understand, manipulate, or summarise human language in forms of text or speech. A key focus lies in a model's ability to capture important information in language using machine learning and rule-based language modelling [1]. Selecting and retaining important parts of an input sequence and understanding context remains a challenge for machines [2].

Automatic text summarisation, a subfield of NLP, involves creating a concise and accurate overview of a text document whilst preserving the critical content and the overall context. Automatic text summarisation has been mostly oriented towards news or science articles, due to the high availability of human annotated data sets [3]. Given the exponential increase in the amount of research published every year, there is an unprecedented volume of diverse textual data which leads to an overwhelming amount of scientific information. Automatic text summarisation has the potential to help reduce this information overload, potentially leading to time and cost savings [4]. Concise summaries make the exponential growth of scientific publications each year more manageable to read [5]. Indeed, there has been growing interest in automatic text summarisation as a method for helping researchers, clinicians and other stakeholders seeking information to efficiently obtain the ''gist'' in a given topic by producing a textual summary from one or multiple document [6].

A common distinction in text summarisation methods is between extractive and abstractive approaches [7, 8]. In extractive text summarisation, phrases of the original text are ranked according to their importance, reordered, and used to create the summary. The ranking is done by assigning relative weights based on the frequency of a word or phrase, or by comparing it to a corpus and determining whether a phrase belongs to a particular domain.

Due to the rise in deep learning methods in recent years, the focus has shifted toward abstractive summarisation [9]. Abstractive text summarisation is more closely aligned with human-created summaries. In contrast to extractive summarisation, abstractive summarisation generates words and phrases that are not present in the original text which are then used for the final summary. This method is more challenging than extractive summarisation methods because it does not guarantee a baseline of grammar and correctness of information. Recent advances have made it possible to generate longer abstract summaries using pre-trained transformer models [10]. A comprehensive review of text summarisation methods has been published previously, including detailed discussion on extractive and abstractive approaches [11], as well as its application to the summarisation of medical content [12].

**Early Generations of NLP methods**

Before the introduction of transformers and inductive transfer learning in NLP, the focus was on stemming (i.e., cutting off the end or the beginning of the word) or lemmatising (i.e., converting a word to its meaningful base form, considering the context) and creating shallow models such as Support Vector Machines. This bore the necessity of training models from scratch for every task, which became increasingly difficult when data for particular tasks were lacking. Moreover, there was a reliance on manually created features rather than learning deep hierarchical representations [13].

Recurrent Neural Networks (RNNs) and Convolutional Neural Networks (CNNs) have transformed the area of machine learning due to their success in image recognition, time-series forecasting, language understanding and other application areas[14, 15]. CNNs are good at extracting features while RNNs are better at dealing with sequential data and retaining temporary input information which is important in language processing for contextualised word embedding. Yin and colleagues [16] compared RNNs and CNNs on a variety of NLP tasks and found that it was not possible to determine which method was superior. However, the authors did not compare these approaches in relation to text summarisation in general.

Traditional RNNs were extended to support gated architectures. Two of the most prominent RNN extensions are Long-Short Term Memory (LSTM) and Gated Recurrent Units (GRUs). Both methods overcome the exploding and vanishing gradient problem which can occur in training RNNs where the gradient can get very small, preventing the weights from updating, or very large when activation functions whose derivatives can get large, are used [17]. Goldberg [18] argues that LSTM and GRU models have provided the most significant contribution to NLP until 2017, as they are adept at capturing key information and retaining it over the long term, which is a vital attribute for language processing. This is also supported by Van Houdt [19] who found that LSTM greatly improved NLP technology embedded in Google Translate, Speech Recognition and Amazon's Alexa. LSTM itself has seen multiple generations of improvements over time, such as the 'forget gate' [20].

To conclude, LSTM models consist of memory blocks, which are able to retain key information long term, however they are still not ideally suited to many NLP tasks due to their sequential nature, which means that words in a sentence are processed word by word rather than as whole sentences as with Transformer models.

**Transformer models**

Transformers were introduced into machine translation with the aim of avoiding recursion to increase parallel computation (to reduce training time) and also to reduce performance drops due to long dependencies. Transformers are non-recurrent sequence-to-sequence tools that combine an encoder with a decoder and usually consist of attention mechanisms to improve context capturing [21]. In principle, self-attention maps the relationships between words at different positions in an input sequence to create a representation and capture the relative importance of tokens compared to others [22]. In the encoder, self-attention calculates an attention score based on the output of the previous layer and compares the score with all other scores of words of the input sequence. The advantages of transformers include easier parallelisability and shorter training time compared to recurrent or convolutional approaches. Self-attention layers are usually stacked multiple times in the encoder and decoder to improve the computation of dependency relationships between words of the same sentence.

Transformers have opened a new chapter in NLP. Some authors even go as far as to call them a revolution[23, 24]. To illustrate, pre-trained transformer models (e.g. Open AI Transformer) were able to achieve 10-20% better results than the previous state of the art in a variety of NLP challenges, such as SQuAD (Stanford Question Answering Dataset) [25, 26], SNLI (Stanford Natural Language Inference Corpus) [27], SRL (Semantic Role Labelling) and SST-5 (Stanford Sentiment Treebank [28]. Ruder [29] calls the emergence of large pre-trained transformers NLP's ImageNet moment because both high- and low-level features are now learnt by models, similar to the transfer-learning approach with ImageNet in the computer vision research community. 'Transfer learning' is based on the human ability to acquire knowledge in one area and use this knowledge to approach a related task. In contrast, earlier NLP techniques were shallower because the features were mainly analysed in the first embedding layer. Transformers are pre-trained on large data sets and can be fine-tuned to solve a variety of NLP tasks, similar to computer vision models which are pre-trained on ImageNet and then fine-tuned to work on different computer vision tasks [30] [31].

**Evaluation metrics**

Metrics are important for model evaluations and comparisons. Popular metrics in machine learning include accuracy, recall, precision, and F1-Score (the harmonic mean of precision and recall). However, in text summarisation, the choice of metrics is not straightforward because generated words, phrases, and whole sentences need to be evaluated instead of labels. One approach is to have humans, ideally subject matter experts, evaluate the generated summaries based on readability, coherence, grammar, and information value. Yet this is time-consuming which renders this approach unfeasible for most cases since developers rely on quick performance evaluation of their models [32].

A good metric should be reliable, consistent yet sensitive to changes in the model or data, and generally applicable to a wide range of tasks [33]. This criterion, however, may be difficult to satisfy with a single metric. Recall-Oriented Understudy for Gisting Evaluation (ROUGE) is a common metric used in text summarisation. ROUGE measures the lexical overlaps between generated summaries and the corresponding full texts. A major limitation of this approach is that it does not give any indication of how meaningful, fact-based, or relevant the generated text bodies are [34, 35]. Moreover, ROUGE tends to assign higher scores to longer summaries [36]. There have been calls for new evaluation metrics needed in the area of NLP [37]. Ng and Abrecht [38] introduced ROUGE-WE, an extension to ROUGE, which considers the cosine similarity during matching. Similarly, BertScore compares generated and original text at the token level while also considering cosine similarity [39].

Bi-Lingual Evaluation Understudy (BLEU) is a precision focused metric for evaluating generated texts, which is mainly used for evaluating machine translation tasks [40]. However, it can also be applied to text summarisation as it compares the closeness of a generated text to a reference text by matching n-grams (i.e., a sequence of N words). BLEU has been criticised for not offering a recall-based metric that would indicate the quality of the produced summary. Subsequently, Banerjee and Lavie [33] introduced the Metric for Evaluation of Translation with Explicit ORdering (METEOR), a metric that was designed to address BLEU's weaknesses. METEOR calculates the harmonic mean (F1) of unigram precision and recall, while weighting recall higher than precision.

In summary, ROUGE, BLUE and METEOR are commonly used evaluation metrics for text summarisation models. Developing metrics for NLP tasks is an ongoing research area, and suitability varies greatly depending on the task and dataset, therefore there are no general recommendations on which metric to use.